\documentclass[journal,letterpaper]{IEEEtran}
\usepackage{amsmath,amsfonts}
\usepackage{algorithmic}
\usepackage{algorithm}
\usepackage{array}
\usepackage[caption=false,font=footnotesize]{subfig}
\usepackage{textcomp}
\usepackage{stfloats}
\usepackage{url}
\usepackage{verbatim}
\usepackage{authblk}
\usepackage{graphicx}
\usepackage[caption=false,font=footnotesize]{subfig}
\usepackage{multirow}
\usepackage{xcolor}
\usepackage{booktabs}
\usepackage{cite} 
\newcommand{\method}{\textsc{DROID}}

\newcommand{\thresholdrule}{thresholded decision rule}

\begin{document}

\title{%DETER: A Dual Encoding and Threshold-Based Re-Classification Framework for Robust Out-of-Scope Intent Classification
DROID: Dual Representation for Out-of-Scope Intent Detection}

\author[1, 2]{Wael Rashwan}
\author[2]{Hossam M. Zawbaa}
\author[3]{Sourav Dutta}
\author[4]{Haytham Assem}
\affil[1] {School of Business, Maynooth University (Ireland), wael.rashwan@mu.ie}
\affil[2]{Technological University Dublin (Ireland),  hossam.zawbaa@gmail.com}
\affil[3]{Huawei Research Centre (Ireland) sourav.dutta2@huawei.com}
\affil[4]{Amazon Alexa AI (United Kingdom). hithsala@amazon.co.uk}

\markboth{Journal of \LaTeX\ Class Files,~Vol.~14, No.~8, August~2021}%
{Shell \MakeLowercase{\textit{et al.}}: A Sample Article Using IEEEtran.cls for IEEE Journals}

\IEEEpubid{0000--0000/00\$00.00~\copyright~2025 IEEE}
% Remember, if you use this, you must call \IEEEpubidadjcol in the second

\maketitle

\begin{abstract}
Detecting out-of-scope (OOS) user utterances remains a key challenge in task-oriented dialogue systems and, more broadly, in open-set intent recognition. Existing approaches often depend on strong distributional assumptions or auxiliary calibration modules. We present \textbf{DROID} (Dual Representation for Out-of-Scope Intent Detection), a compact end-to-end framework that combines two complementary encoders—the Universal Sentence Encoder (USE) for broad semantic generalization and a domain-adapted Transformer-based Denoising Autoencoder (TSDAE) for domain-specific contextual distinctions. Their fused representations are processed by a lightweight branched classifier with a single calibrated threshold that separates in-domain and OOS intents without post-hoc scoring. To enhance boundary learning under limited supervision, DROID incorporates both synthetic and open-domain outlier augmentation. Despite using only 1.5M trainable parameters, DROID consistently outperforms recent state-of-the-art baselines across multiple intent benchmarks, achieving macro-F1 improvements of \textbf{6--15\%} for known and \textbf{8--20\%} for OOS intents, with the largest gains in low-resource settings. These results demonstrate that dual-encoder representations with simple calibration can yield robust, scalable, and reliable OOS detection for neural dialogue systems.
\end{abstract}


\section{Introduction}

\IEEEPARstart{C}{onversational} AI systems are a primary interface for user assistance across sectors such as customer service, healthcare, and finance. A core requirement is \emph{intent classification}—mapping utterances to predefined intents so downstream components can act appropriately \cite{Coucke_2018}. Equally critical is detecting \emph{out-of-scope (OOS)} utterances—inputs that do not belong to any trained intent—because misrouting unknowns degrades user experience and safety \cite{Cavalin_2020}. This challenge is amplified in low-data, domain-specific deployments where curated intent coverage is inherently incomplete \cite{Casanueva_2020}. We therefore cast the problem as \emph{open-set recognition for text}, wherein a model must confidently assign \emph{in-domain} intents while rejecting OOS inputs.

Despite substantial gains from pretrained transformers in intent classification \cite{devlin-etal-2019-bert}, OOS detection remains difficult. Confidence-based heuristics (e.g., maximum softmax probability) are brittle and sensitive to calibration \cite{hendrycks2018baseline}; open-set extensions such as OpenMax still rely on parametric tail assumptions \cite{bendale2015open}. Density- and feature-space approaches (e.g., LOF, Mahalanobis-based detectors) can suffer under feature collapse or domain shift \cite{Breunig2000,xu-etal-2021-unsupervised}. Boundary/point methods (e.g., DOC, ARPL, ADB/DA-ADB) improve separability but often introduce complex objectives or adversarial components \cite{shu-etal-2017-doc, Guangyao2022, Zhang_2021 ,DA-ADB_2023}. Synthetic outlier augmentation has proven effective by casting training as a $(K{+}1)$-class problem that mixes feature-space constructs with open-domain negatives \cite{Zhan_2021}. Recent advances further refine representation learning and boundaries (e.g., TCAB; autoencoder-regularized fine-tuning) \cite{chen2024tcab,zhang2024finetuning}, leverage class-name semantics (SCOOS) \cite{gautam2024scoos}, or jointly shape clusters and adaptive boundaries (CLAB) \cite{liu2023clab}. Complementary interactive approaches formulate post-hoc clarification for uncertain predictions \cite{hengst2024cicc}.

Large language models (LLMs) provide compelling zero/few-shot baselines via prompting or instruction tuning \cite{marzagao-etal-2024-beyond,zhang-etal-2024-discrimination,arora-varma-2024-intent}. However, their inference latency and computational cost limit real-time deployment, and recent evidence shows smaller, well-adapted models can remain competitive for open-intent settings \cite{arora-varma-2024-intent,marzagao-etal-2024-beyond}.

We introduce \textbf{DROID} (\emph{Dual Representation for Out-of-Scope Intent Detection}), an efficient end-to-end framework that addresses these limitations. DROID integrates two complementary sentence encoders—the \emph{Universal Sentence Encoder (USE)} for broad semantic coverage \cite{Daniel_2018} and a domain-adapted \emph{Transformer-based Denoising AutoEncoder (TSDAE)} for fine-grained, task-specific nuance \cite{Wang_2021}—within a light-weight branched classifier. A \emph{single calibrated threshold} on the softmax outputs \textbf{(tuned only on ID validation data)} separates known from OOS intents at inference, avoiding post-hoc detectors and strong distributional assumptions. To further enhance robustness, especially under few-shot label regimes, DROID trains with \emph{synthetic feature-space outliers} and \emph{open-domain negatives}, \textbf{building on} \cite{Zhan_2021}, but within a dual-representation pipeline.

\textbf{Contributions.}
\begin{itemize}
    \item \emph{Dual-encoder representation.} We fuse USE \cite{Daniel_2018} with a \emph{domain-adapted} TSDAE \cite{Wang_2021} to construct richer, more discriminative utterance embeddings for open-intent recognition, complementing recent representation-learning advances \cite{chen2024tcab,zhang2024finetuning}.
    \item \emph{Thresholded decision rule.} A single calibrated threshold (tuned only on ID validation data). It separates known from OOS intents at inference, avoiding post-hoc detectors and strong distributional assumptions.
    \item \emph{Outlier augmentation.} We combine \emph{synthetic} feature-space outliers with \emph{open-domain} negatives, extending mixture-based open-class training \cite{Zhan_2021} in a dual-representation setting.
    \item \emph{Efficiency and scalability.} The trainable part of DROID comprises \textbf{1.5M} parameters—far smaller than common open-set baselines \cite{Zhan_2021}—supporting real-time and resource-constrained use.
    \item \emph{Extensive validation.} On \textbf{CLINC-150} \cite{Larson_2019}, \textbf{BANKING77} \cite{Casanueva_2020}, and \textbf{STACKOVERFLOW} \cite{Xu_2015}, DROID achieves consistent gains, with macro-F1 improvements of up to \textbf{16\%} (known) and \textbf{24\%} (unknown) over strong baselines, and remains robust under few-shot label ratios with analyses of thresholding, class weighting, encoding, and outlier quantities.
\end{itemize}

\noindent\textbf{Relation to prior work.}
This manuscript substantially extends our conference version, \emph{DETER} \cite{zawbaa2024deter}. For clarity, we refer to the extended framework as \emph{DROID} throughout. Compared to \cite{zawbaa2024deter}, we (i) fully specify the architecture (layer sizes, normalization, dropout), (ii) add comprehensive ablations on the \thresholdrule, class weighting, and outlier composition/quantity as well as encoding strategies, (iii) deepen the analysis of domain adaptation effects (USE+TSDAE), and (iv) expand experimental validation across multiple known-intent and label ratios with 10 seeds for statistical robustness, including few-shot behavior and error analysis. On \textbf{CLINC-150}, \textbf{BANKING77}, and \textbf{STACKOVERFLOW}, DROID achieves macro-F1 gains of up to \textbf{16\%} (known) and \textbf{24\%} (unknown) over strong baselines.

\noindent\textbf{Paper outline.} Section~\ref{sec:related_literature} reviews related work; Section~\ref{sec:droid} formalizes the problem and presents \method; Section~\ref{sec:experimental_setup} details datasets, baselines, and implementation; Section~\ref{sec:results} reports results and ablations; Section~\ref{sec:discussion} discusses findings and limitations; Section~\ref{sec:conclusion} concludes.

% put this once in your preamble (or just before Related Work)
\newcommand{\preprint}{\textit{(preprint)}}

\section{Related Literature}
\label{sec:related_literature}

\noindent\textbf{Terminology.}
We study \emph{out-of-scope (OOS)} intent detection as \emph{open-set recognition (OSR)} for text: a model must confidently assign in-domain (known) intents and reject OOS inputs. We focus on \emph{NLP dialogue/intent} methods and exclude vision-first OOD/OSR baselines.

\subsection{Representation Learning}
Sentence-level representations are central to open-intent recognition \cite{Nils_2019}. While pretrained transformers boost intent classification \cite{devlin-etal-2019-bert}, ID/OOS separability often requires task-specific adaptation \cite{Vibhav_2021}. Recent work sharpens separability via structured objectives, e.g., triplet-contrastive learning with adaptive boundaries (TCAB) \cite{chen2024tcab} and autoencoder-regularized fine-tuning \cite{zhang2024finetuning}\preprint{}. Our approach complements these directions by pairing two \emph{complementary} encoders (USE and a domain-adapted TSDAE) while keeping the decision mechanism simple.

\subsection{Paradigms for OOS/OSR in Intent Detection}

\paragraph{Density/feature-space (NLP).}
Transformer-based Mahalanobis features (MDF) aggregate layer-wise distances and use a one-class SVM for OOS on intent corpora \cite{xu-etal-2021-unsupervised}. DeepUnk learns margin-enhanced features and applies LOF post-hoc \cite{lin-xu-2019-deep}, and SEG uses large-margin Gaussian mixture embeddings with LOF for detection \cite{yan-etal-2020-unknown}. KNNCL leverages KNN-guided contrastive learning to compact intent clusters \cite{zhou-etal-2022-knn}. \noindent A succinct comparison of representative NLP intent/OOS methods is provided in Table~\ref{tab:nlp_oos_dialogue}.

\paragraph{Boundary/point/semantics (NLP).}
DOC introduces one-vs-rest sigmoids with per-class thresholds \cite{shu-etal-2017-doc}. Distance-aware adaptive boundaries (ADB/DA-ADB) learn per-class margins for open-intent classification \cite{Zhang_2021, DA-ADB_2023}. SCOOS tightens decision regions using class-name semantics (BERT) with an SVAE prior \cite{gautam2024scoos}. CLAB couples K-center contrastive clustering with adaptive boundary scaling \cite{liu2023clab}. TCAB jointly optimizes contrastive structure and a boundary \cite{chen2024tcab}. These approaches improve separability but often add objectives or auxiliary heads.

\paragraph{Synthetic outlier augmentation (NLP).}
$(K{+}1)$ training with outliers is effective for intents.Zang~et~ al.~\cite{Zhan_2021} synthesize \emph{feature-space} outliers via convex combinations of representations from distinct known intents and mix them with \emph{open-domain negatives}, training a unified $(K{+}1)$ classifier.
 
\paragraph{Dynamic/interactive (NLP).}
AIDOIL integrates anchors for dynamic matching to represent diverse OOS without significant augmentation \cite{yin2025aidoil}. CICC converts classifier uncertainty into clarification questions with statistical coverage guarantees \cite{hengst2024cicc}, trading rejections for interaction.

\subsection{Modern Training Strategies and the Role of LLMs}
LLMs enable zero/few-shot intent/OOS via prompting or instruction tuning \cite{marzagao-etal-2024-beyond,zhang-etal-2024-discrimination,arora-varma-2024-intent}, but latency/memory often preclude real-time routing; smaller, well-adapted models remain competitive in open-intent settings \cite{arora-varma-2024-intent,marzagao-etal-2024-beyond}. Parameter-efficient tuning and distillation mitigate costs, yet many deployments still prefer compact classifiers with predictable calibration.

\subsection{Our Contribution in Context}
\textbf{DROID} couples two complementary encoders (USE \cite{Daniel_2018} and a domain-adapted TSDAE \cite{Wang_2021}) with a \emph{single calibrated threshold} and mixed outlier augmentation \cite{Zhan_2021}, avoiding adversarial training and auxiliary scoring heads while retaining a small trainable footprint.

\begin{table*}[t]
\centering
\caption{Representative NLP methods for OOS detection in task-oriented dialogue/intent classification. “OOS data?” denotes whether training uses synthetic or real OOS.}
\label{tab:nlp_oos_dialogue}
\renewcommand{\arraystretch}{.8}
\setlength{\tabcolsep}{2.0pt}
\footnotesize
\begin{tabular}{@{}l l c p{5.3cm} p{2cm}@{}}
\toprule
\textbf{Family} & \textbf{Method} & \textbf{OOS data?} & \textbf{Core idea (intent setting)} & \textbf{Extra module} \\
\midrule
Baseline (NLP) 
& MSP \cite{hendrycks2018baseline} & No & Max softmax as confidence; reject below threshold on intent datasets. & -- \\
\midrule
Density/feature (NLP)
& MDF (transformers) \cite{xu-etal-2021-unsupervised} & No & Layer-wise transformer features; Mahalanobis distances aggregated, then one-class SVM for OOS. & One-class SVM \\
& DeepUnk \cite{lin-xu-2019-deep} & No & Learn margin-enhanced features for intents; LOF post-hoc to flag OOS utterances. & LOF \\
& SEG \cite{yan-etal-2020-unknown} & No & Large-margin Gaussian mixture embeddings for intents; LOF for OOS detection. & LOF \\
& KNNCL \cite{zhou-etal-2022-knn} & No & KNN-guided contrastive learning to compact intent clusters and expose OOS. & -- \\
\midrule
Boundary/semantics (NLP)
& DOC \cite{shu-etal-2017-doc} & No & One-vs-rest sigmoids with per-class thresholds for open-intent recognition. & Thresholds \\
& ADB / DA-ADB \cite{Zhang_2021,DA-ADB_2023} & No & Distance-aware features with adaptive per-class boundaries for open-intent classification. & Boundary head \\
& SCOOS \cite{gautam2024scoos} & No & Class-name semantics (BERT) + SVAE prior tighten known-intent regions; depends on label quality. & SVAE head \\
& CLAB \cite{liu2023clab} & No & K-center contrastive clustering + adaptive boundary scaling for intent spaces. & Boundary scaler \\
& TCAB \cite{chen2024tcab} & No & Triplet-contrastive learning with adaptive boundary to separate known/unknown intents. & Boundary term \\
\midrule
(K$+$1) with outliers (NLP)
& (K$+$1)-way \cite{Zhan_2021} & Synthetic+Open & Unified $(K{+}1)$ classifier trained with convex-combo \emph{synthetic} outliers + \emph{open-domain negatives}. & -- \\
\midrule
Dynamic/interactive (NLP)
& AIDOIL \cite{yin2025aidoil} & No (anchors learned) & Anchor-integrated dynamic matching to represent diverse OOS without large augmentation. & Anchor memory \\
& CICC (interactive) \cite{hengst2024cicc} & No & Converts classifier uncertainty into clarification questions with coverage guarantees. & Clarification module \\
\midrule
LLM-based (NLP)
& Prompt/IT baselines \cite{marzagao-etal-2024-beyond,zhang-etal-2024-discrimination,arora-varma-2024-intent} & No & Zero/few-shot intent/OOS via prompting or instruction tuning; strong but higher latency/cost. & -- \\
\midrule
\textbf{Dual encoders + threshold (ours)}
& \textbf{DROID} & Synthetic+Open & USE + domain-adapted TSDAE fused; single calibrated threshold (ID-only validation) separates known vs.~OOS; compact head. & -- \\
\bottomrule
\end{tabular}
\end{table*}

\subsection{Modern Training Strategies and the LLM Revolution}

\label{ssec:llm_revolution}

Training strategies, particularly for data-scarce scenarios, have become as crucial as model architecture itself. While earlier work effectively used data augmentation with synthetic outliers \cite{Zhan_2021}, the field is now grappling with the paradigm shift introduced by \textbf{Large Language Models (LLMs)}. The immense world knowledge and generative power of models like GPT-4 and Llama have reshaped the landscape, moving the focus from purely discriminative classifiers to generative solutions.

The primary way LLMs have been leveraged is through prompt-based inference, which requires no task-specific training. A significant line of current research involves using these prompt-based strategies for few-shot or zero-shot detection \cite{ marzagao-etal-2024-beyond, arora-varma-2024-intent} and systematically investigating LLM performance on out-of-domain intents to understand their true capabilities and failure points \cite{marzagao-etal-2024-beyond}.

To move beyond the limitations of simple prompting, more advanced adaptation techniques are being explored. Instruction tuning, for instance, has emerged as a powerful method for refining LLM behavior for specific tasks. This approach reformulates intent detection as a generative task, proving especially effective in challenging low-resource scenarios \cite{zhang-etal-2024-discrimination}.

However, the adoption of LLMs is not without significant challenges. Their immense computational cost and high inference latency make them impractical for many real-time applications. Furthermore, recent "reality check" investigations have shown that smaller, efficiently fine-tuned models can still outperform these large-scale counterparts in specific contexts, highlighting a critical trade-off between generative power and practical viability \cite{arora-varma-2024-intent}. This ongoing tension between massive, general-purpose LLMs and smaller, specialized models defines the current research frontier. A consolidated view of these approaches—their supervision, assumptions, and deployment needs—appears in Table~\ref{tab:nlp_oos_dialogue}.

\noindent\textbf{Positioning.}
Prior intent/OOS methods typically trade off simplicity and robustness: representation/boundary refinements add training complexity \cite{chen2024tcab,zhang2024finetuning}, semantic-guided models depend on label quality \cite{gautam2024scoos}, and several approaches introduce auxiliary scores or adversarial components. In contrast, \textbf{DROID} uses two complementary encoders (USE for broad semantics and a \emph{domain-adapted} TSDAE for task nuance) fused by a small branched head, a \emph{single} thresholded decision rule calibrated \emph{only} on in-domain validation data (no labeled OOS), and mixed outlier augmentation in a $(K{+}1)$ setup \cite{Zhan_2021}. This design avoids parametric tail assumptions and post-hoc detectors while keeping the trainable footprint small; detailed results appear in Sec.~\ref{sec:results}.

%--------------

\section{DROID: Dual Encoders with a Thresholded $(K{+}1)$ Classifier}
\label{sec:droid}

\subsection{Problem Setup}
\begin{figure*}[t]
\centering
\includegraphics[width=.8\textwidth]{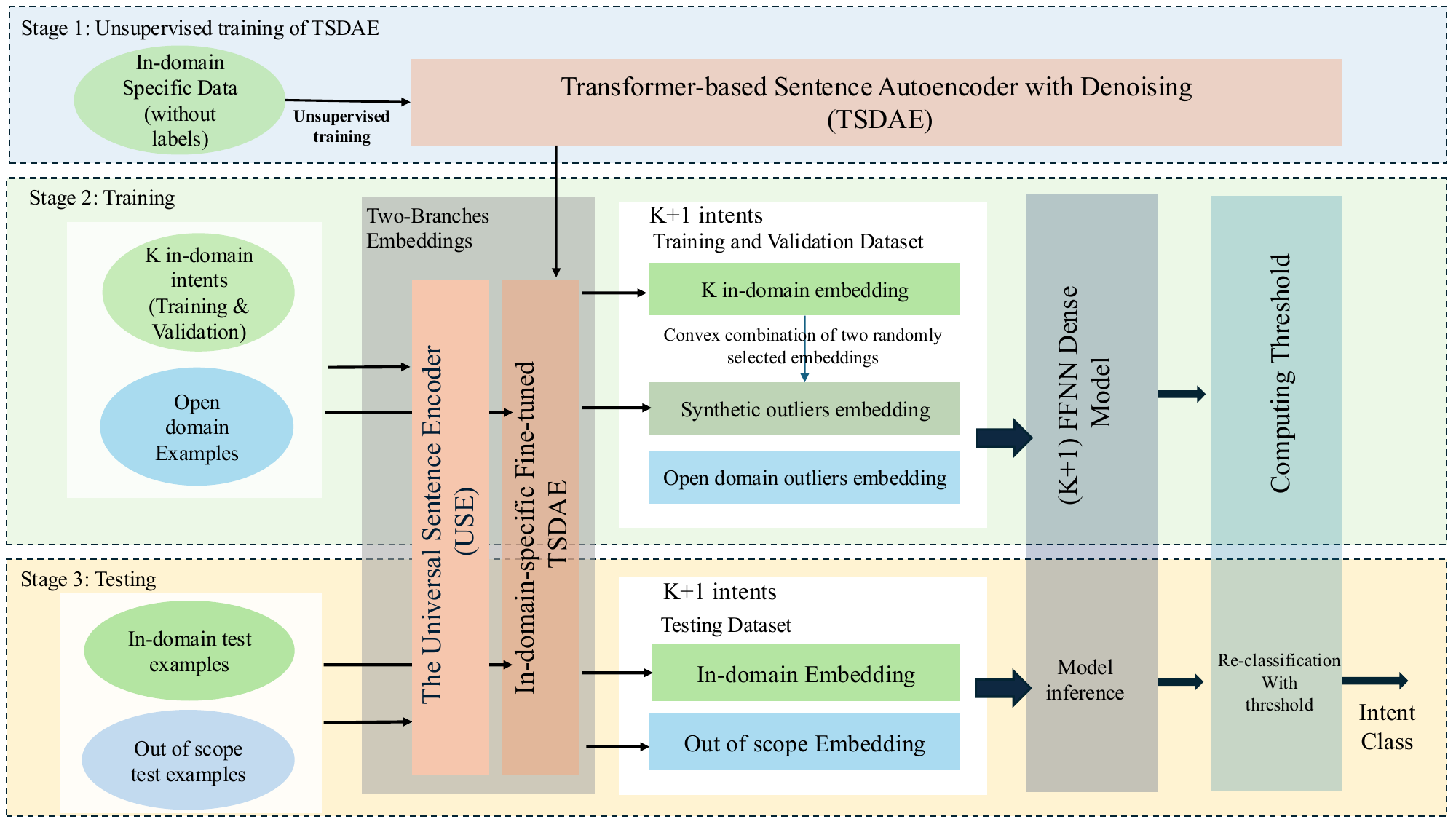}
\caption{End-to-end pipeline of \textbf{DROID}. \emph{Stage 1:} unsupervised domain adaptation of the TSDAE encoder on in-domain unlabeled text. \emph{Stage 2:} supervised $(K{+}1)$ training with two frozen encoders (USE, domain-adapted TSDAE); per-branch projections produce embeddings for known intents, \emph{synthetic} feature-space outliers (convex mixes of distinct known-class embeddings), and \emph{open-domain} negatives. A light-weight MLP learns the $(K{+}1)$ classifier, and a single threshold $T$ is calibrated on in-domain validation data. \emph{Stage 3:} inference on test utterances using the thresholded decision rule to separate known vs.\ OOS.}
\label{fig:droid_overview}
\end{figure*}

Let $S_{\text{known}}=\{C_1,\ldots, C_K\}$ be the set of in-domain (known) intents and let $C_{K+1}=\text{OOS}$ denote the reject class. Given an utterance $u$, DROID is trained as a $(K{+}1)$-way classifier producing $p(c\mid u)\in\mathbb{R}^{K+1}$. At inference, we apply a single \emph{thresholded decision rule}:
\begin{equation}
\label{eq:decision}
\hat{c}(u)=
\begin{cases}
\arg\max_i\, p(c{=}C_i\mid u), & \text{if } \max_i\, p(c{=}C_i\mid u)\ge T,\\[2pt]
\text{OOS}, & \text{otherwise}.
\end{cases}
\end{equation}
Assumptions: (i) no labeled OOS data is used for threshold calibration; (ii) encoders are frozen during supervised training; (iii) synthetic and open-domain samples are labeled as OOS during training only. The overall pipeline is illustrated in Fig.~\ref{fig:droid_overview}.

\subsection{Sentence Encoders}
\textbf{Universal Sentence Encoder (USE).} We use the Transformer-based USE (TF-Hub), mapping $u\mapsto E_{\text{USE}}(u)\in\mathbb{R}^{512}$. USE parameters remain \emph{frozen} throughout.

\noindent\textbf{TSDAE (domain-adapted).} We train a RoBERTa-based TSDAE via denoising on unlabeled \emph{target-domain} text following \cite{Wang_2021}. For $u$ and a corrupted $\tilde{u}$ (token deletion/masking), TSDAE minimizes

\begin{equation}
\label{eq:tsdae}
L_{\text{TSDAE}}
= 1 - \cos\!\big(E_{\text{TSDAE}}(u),\,E_{\text{TSDAE}}(\tilde{u})\big).
\end{equation}
\noindent\textit{where } $E_{\text{TSDAE}}(u)\in\mathbb{R}^{768}$.

Unless stated otherwise, TSDAE is \emph{frozen} during supervised training (Sec .~\ref {sec:results} ablates adaptation sources and freezing).

\subsection{Outlier Construction}
We enrich the OOS signal at training time with two sources.

\noindent\textbf{Synthetic (feature-space) outliers.} Following \cite{Zhan_2021}, we synthesize hard OOS by convexly mixing representations from two \emph{distinct} known classes. Let $h_\alpha,h_\beta$ be representation vectors sampled from different classes; we form
\begin{equation}
\label{eq:synthetic}
h^{\text{OOS}}=\theta\, h_\beta + (1-\theta)\, h_\alpha,\quad \theta\sim U(0,1).
\end{equation}
Unless specified, we generate $h_\alpha,h_\beta$ in the \emph{fused} space defined in \eqref{eq:fuse} (post-branch projection), which empirically yields diverse but on-manifold negatives. All synthetic samples are labeled OOS.

\noindent\textbf{Open-domain negatives.} We add generic negatives from SQuAD~2.0 question text \cite{Rajpurkar_2018} (length filter $5{\le}|u|{\le}64$, de-duplication), following \cite{Zhan_2021}. These are encoded by USE/TSDAE and treated as OOS during training.

\noindent\textbf{Quantities and mixing.} Per epoch, we sample comparable counts of synthetic and open-domain OOS; unless otherwise stated,, we use $N_{\text{syn}}{=}500$ and $N_{\text{open}}{=}500$ per epoch. In mini-batches, we maintain an OOS fraction between $20\%$ and $40\%$ (tuned in Sec .~\ref{sec:results}).

\subsection{Architecture}
\begin{figure}[t]
\centering
\includegraphics[width=.4\textwidth]{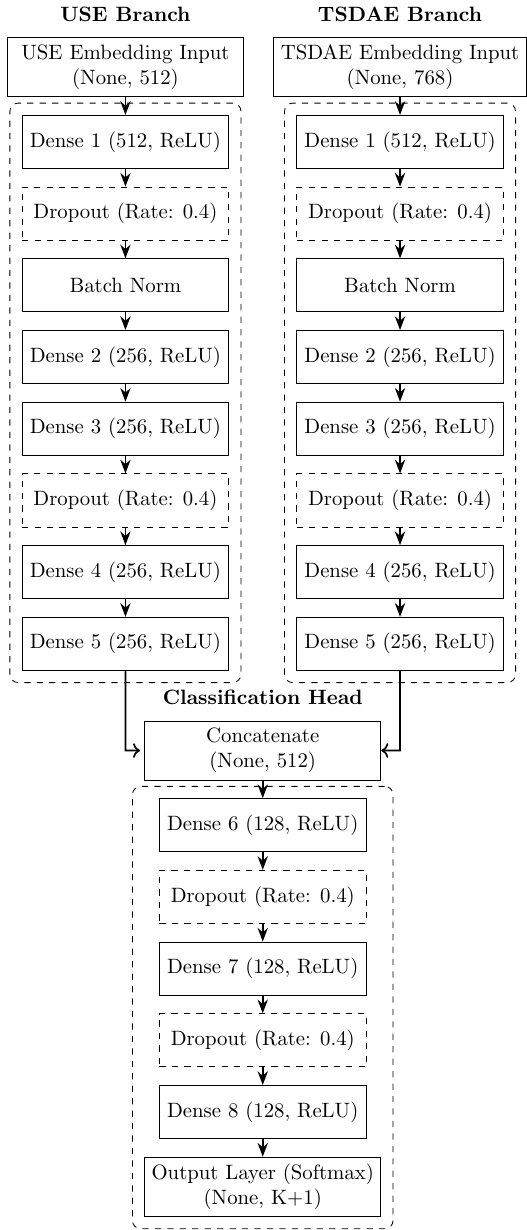}
\caption{Head architecture of \textbf{DROID}. USE and TSDAE embeddings are passed through parallel 5-layer MLP branches (sizes $512$–$256$–$256$–$256$–$256$; ReLU; BatchNorm and Dropout $0.4$ after the first two layers) to 256-d projections. The projections are concatenated (512-d) and fed to a 3-layer classifier (sizes $128$–$128$–$128$; ReLU; Dropout $0.4$ after the first two layers) followed by a linear layer to $(K{+}1)$ logits and softmax. A single calibrated threshold on the maximum softmax decides OOS at inference.}
\label{fig:droid_arch}
\end{figure}

Figure~\ref{fig:droid_overview} summarizes DROID; Fig.~\ref{fig:droid_arch} details the head and per-encoder branches.
 Two per-encoder branches project embeddings into a common space; features are fused and classified by a small MLP head.

\noindent\textbf{Per-encoder branches.} For encoder $e\in\{\text{USE},\text{TSDAE}\}$:
\begin{align}
h'_e &= f_e\!\big(E_e(u)\big), \quad h'_e\in\mathbb{R}^{256}, \\
f_e &: \text{MLP with hidden sizes } (512,256,256,256,256),
\end{align}
with ReLU activations and BatchNorm+Dropout ($p{=}0.4$) after the first two layers. (Weights are trainable; encoders are frozen.)

\noindent\textbf{Fusion and classifier.}
The fusion and $(K{+}1)$ classifier are depicted on the right side of Fig.~\ref{fig:droid_arch}.

\begin{equation}
\label{eq:fuse}
h_{\text{fused}}(u)=\big[h'_{\text{USE}};h'_{\text{TSDAE}}\big]\in\mathbb{R}^{512}.
\end{equation}
A 3-layer MLP head with hidden sizes $(128,128,128)$ (ReLU; Dropout $0.4$ after first two) maps to logits $z\in\mathbb{R}^{K+1}$:
\begin{equation}
\label{eq:softmax}
z=W_{\text{final}}\,f_{\text{head}}\!\big(h_{\text{fused}}(u)\big)+b_{\text{final}},\qquad
p(c\mid u)=\mathrm{softmax}(z).
\end{equation}
With encoders frozen, the trainable head (branches $+$ classifier) has $\sim$1.56M parameters.

\subsection{Training Objective and Regularization}
Given a batch $\{(u^{(j)},y^{(j)})\}_{j=1}^M$ with one-hot $y^{(j)}\in\{e_1,\ldots,e_{K+1}\}$, we minimize
\begin{equation}
\label{eq:ce}
L_{\text{CE}}=-\frac{1}{M}\sum_{j=1}^{M}\sum_{i=1}^{K+1} w_i\, y_i^{(j)}\log p_i(c\mid u^{(j)}),
\end{equation}
where $w_i$ are class weights to mitigate imbalance between known classes and OOS. Early stopping on validation accuracy is applied; optimizer and schedules are detailed in Sec.~\ref {sec:experimental_setup}. Unless stated, we do not fine-tune encoder parameters.

\subsection{Threshold Calibration}
We select a single threshold $T\in[0,1]$ using \emph{in-domain validation data only} (no labeled OOS), to avoid leakage:
\begin{itemize}
\item Compute scores $s(u)=\max_i p(c{=}C_i\mid u)$ for validation utterances from $S_{\text{known}}$.
\item Either (i) choose $T$ as the $(1{-}\alpha)$-quantile of $\{s(u)\}$ for a target ID coverage $1{-}\alpha$, or (ii) grid-search $T\in\{0.00,0.02,\ldots,1.00\}$ to maximize validation accuracy on known intents.\footnote{Sec.~\ref{sec:results} shows both strategies give similar operating points; we report the grid-search variant unless otherwise noted.}
\end{itemize}
At test time, we apply \eqref{eq:decision}. We additionally report a proxy-OOS calibration (held-out intents as unknowns) in ablations to illustrate sensitivity (Sec .~\ref{sec:results}).

\subsection{Complexity and Deployment Considerations}
\textbf{Compute.} Inference requires two encoder forward passes plus a small MLP head; with frozen encoders and a 1.56M-parameter head, latency is dominated by encoders. \textbf{Memory.} Only head parameters are trainable; encoders are loaded once and shared across tasks. \textbf{Stability.} Using a single scalar $T$ avoids classwise threshold tuning and post-hoc detectors; we find $T$ is stable across seeds and label ratios (Sec.~\ref{sec:results}). \textbf{Portability.} Because $T$ is calibrated on ID validation only, the procedure does not require labeled OOS for a new domain.

\section{Experimental Setup}
\label{sec:experimental_setup}
We evaluate \textbf{DROID} on standard intent benchmarks and compare against representative open intent detection methods. The setup specifies datasets/splits, training-time outlier usage, encoders, baselines, implementation, threshold calibration, and metrics.

\subsection{Datasets and Protocol}
We use \textbf{CLINC-150} \cite{Larson_2019}, \textbf{BANKING77} \cite{Casanueva_2020}, and \textbf{STACKOVERFLOW} \cite{Xu_2015}. For each dataset, we sample known-intent subsets at \{25\%, 50\%, 75\%\} of the classes; the remaining courses are held out and treated as \emph{unknown} (unselected intents, UI) at test time. We repeat each configuration with fixed random seeds $0$–$9$, following TEXTOIR \cite{zhang-etal-2021-textoir}. Consistent with prior work, the \emph{external} CLINC-150 OOS set (1{,}200 utterances) is always included in the test set, regardless of the primary data set. Table \ref{Table:datasets_details} provides statistics on the data sets in the 25\% known intent ratio.

\begin{table*}[!t]
\caption{Dataset statistics at the 25\% known-intent ratio. ``Unknown'' counts reflect unselected intents (UI), synthetic outliers ($N_{\text{outlier}}$), and Open-Domain (OD) or Out-of-Scope (OOS) samples, as applicable.}
\label{Table:datasets_details}
\centering
\begin{tabular*}{\textwidth}{@{\extracolsep{\fill}} l l ll ll ll @{}}
\toprule
\multirow{2}{*}{\textbf{Dataset}} & \multirow{2}{*}{\textbf{Intent Type}} & \multicolumn{2}{c}{\textbf{Training}} & \multicolumn{2}{c}{\textbf{Validation}} & \multicolumn{2}{c}{\textbf{Testing}} \\
\cmidrule(lr){3-4} \cmidrule(lr){5-6} \cmidrule(lr){7-8}
& & \textbf{Total} & \textbf{25\% Known} & \textbf{Total} & \textbf{25\% Known} & \textbf{Total} & \textbf{25\% Known} \\
\midrule
\multirow{2}{*}{CLINC-150} 
& Known   & 15{,}000 & 3{,}800 & 3{,}000 & 760  & 4{,}500 & 1{,}140 \\
& Unknown & 0        & 11{,}200 (UI) + $N_{\text{outlier}}$ & 0 & 221 (OD) & 1{,}200 & 1{,}200 (OOS) + 3{,}360 (UI) \\
\multirow{2}{*}{BANKING77} 
& Known   & 9{,}003 & 2{,}119 & 1{,}000 & 234 & 3{,}080 & 760 \\
& Unknown & 0      & 6{,}884 (UI) + $N_{\text{outlier}}$ & 0 & 221 (OD) & 1{,}200 & 1{,}200 (OOS) + 2{,}320 (UI) \\
\multirow{2}{*}{STACKOVERFLOW} 
& Known   & 12{,}000 & 3{,}000 & 2{,}000 & 500 & 6{,}000 & 1{,}500 \\
& Unknown & 0       & 9{,}000 (UI) + $N_{\text{outlier}}$ & 0 & 221 (OD) & 1{,}200 & 1{,}200 (OOS) + 4{,}500 (UI) \\
\bottomrule
\end{tabular*}
\hfill\par\smallskip
\noindent\emph{Note:} UI = Unselected (held-out) intents; OOS = external CLINC-150 OOS; OD = Open-Domain negatives used on validation for threshold calibration.
\end{table*}

\subsection{Encoders and Training-Time Outliers}
We use the Transformer-based \textbf{USE} and a \textbf{BERT}-based \textbf{TSDAE} \cite{Wang_2021} (both frozen during supervised training). TSDAE is domain-adapted once via denoising on unlabeled in-domain text and reused across datasets unless otherwise stated.

\subsection{Synthetic Outlier Generation}
\label{sec:synthetic_outliers}

To enhance the model’s ability to delineate known and unknown intent boundaries,
we employ a synthetic outlier generation mechanism inspired by convex feature-space interpolation.
Using the fused representations defined in Eq.~\eqref{eq:fuse}, we randomly sample
two embeddings, $h_\alpha$ and $h_\beta \in \mathbb{R}^{d}$, from distinct known intent classes and interpolate between them to create pseudo OOS examples:
\begin{equation}
\label{eq:synth}
h^{\text{OOS}} = \theta\, h_\beta + (1-\theta)\, h_\alpha, \qquad
\theta \sim U(0,1).
\end{equation}
Such convex interpolations generate samples that lie outside the convex hull of
in-domain clusters, effectively populating low-density regions between intent manifolds.
These “hard” negatives encourage the model to learn compact and well-separated decision
boundaries. Synthetic samples generated are mixed with open domain negatives to form a balanced OOS training set, where the relative proportions of synthetic and OD examples are tuned empirically (typically $1{:}1$) to maintain calibration stability.

Unlike prior OOS augmentation strategies that depend on large external corpora or
adversarial perturbations, this approach is self-contained and computationally efficient, requiring only in-domain data and encoder-derived embeddings. Consequently, it provides a scalable mechanism for strengthening boundary learning in low-resource or privacy-sensitive dialogue applications.

\paragraph*{Open-Domain (OD) Negatives.} In parallel, following the protocol of~\cite{Zhan_2021}, we adopt the SQuAD~2.0 corpus~\cite{Rajpurkar_2018}
as a source of OD utterances, providing linguistically diverse yet semantically unrelated examples.
During tuning, the number of OD and synthetic outliers was varied over $[50,\,4000]$ and $[50,\,16000]$,
respectively. Empirically, a balanced configuration of approximately $500$ OD and $500$ synthetic
samples per epoch yielded the most stable performance across datasets, offering sufficient variety
without over-saturating the training distribution.

\subsection{Baselines}
We compare against representative intent/OOS methods from TEXTOIR \cite{zhang-etal-2021-textoir}: MSP \cite{hendrycks2018baseline}, DOC \cite{shu-etal-2017-doc}, OpenMax \cite{bendale2015open}, LOF \cite{Breunig2000}, DeepUnk \cite{lin-xu-2019-deep}, SEG \cite{yan-etal-2020-unknown}, MDF \cite{xu-etal-2021-unsupervised}, KNNCL \cite{zhou-etal-2022-knn}, ARPL \cite{Guangyao2022}, ADB \cite{Zhang_2021}, DA-ADB \cite{DA-ADB_2023}, and $(K{+}1)$-Way with synthetic+open outliers \cite{Zhan_2021}. For fairness with prior reports, baselines use \texttt{bert-base-uncased} in TEXTOIR.

\subsection{Implementation Details}
Models are implemented in Keras/TensorFlow. We use AdamW \cite{Adam} with categorical cross-entropy, batch size 200, learning rate $10^{-3}$, and a maximum of 1000 epochs with early stopping (patience 100) on validation accuracy. Maximum sequence length is 512 tokens. Unless otherwise noted, we use the operating point of 500 OD and 500 synthetic outliers per epoch.

\subsection{Threshold Calibration}
We select a single threshold \(T\in\{0.00,0.02,\ldots,1.00\}\) on the \emph{validation} set, which contains ID examples (from the selected known intents) and OD negatives (cf. Table~\ref{Table:datasets_details}). We compute \(s(u)=\max_i p(c{=}C_i\mid u)\) and choose \(T\) maximizing validation accuracy for known vs.\ unknown discrimination; the best \(T\) was \(0.7\) in our runs.

\subsection{Metrics}
We report macro F1 for: (i) \textbf{Known} (over the $K$ in-domain classes), (ii) \textbf{Unknown} (the OOS class), and (iii) overall $(K{+}1)$. Let $P$ and $R$ be macro precision/recall over $K{+}1$ classes; then
\begin{equation}
\label{eq:f1}
F1=\frac{2PR}{P+R}.
\end{equation}
Per-class precision/recall are $P_{C_i}=\frac{TP_{C_i}}{TP_{C_i}+FP_{C_i}}$ and $R_{C_i}=\frac{TP_{C_i}}{TP_{C_i}+FN_{C_i}}$. Known-only macro-averages use $i\in\{1,\ldots,K\}$; the OOS F1 uses $i=K{+}1$. We present the aggregate results in the tables; dispersion measures are included only where explicitly stated.

%--------------------
\section{Experimental Results \& Ablation}
\label{sec:results}

We evaluate \method\ on three benchmark intent datasets---\textsc{CLINC-150} \cite{Larson_2019}, \textsc{BANKING-77} \cite{Casanueva_2020}, and \textsc{STACKOVERFLOW} \cite{Xu_2015}---against representative OOS intent detection baselines from TEXTOIR \cite{zhang-etal-2021-textoir}. Unless stated otherwise, we report macro~F1 for (i) in-domain \emph{Known} classes and (ii) the \emph{Unknown} (OOS) class under three coverage regimes, where $25\%$, $50\%$, or $75\%$ of intents are treated as Known during training. Quantitative comparisons at label ratio $1.0$ appear in Table~\ref{Table:full-labeles-results}. Trends under varying label ratios are summarized in Fig.~\ref{fig_FS_All}. Component-wise analyses are provided in Figs.~\ref{fig__threshold}--\ref{fig_open-synth-data}. 

\subsection{Main Comparative Results}
Table~\ref{Table:full-labeles-results} shows that \method\ attains the best mean macro~F1 on both Known and Unknown classes across all datasets and intent-coverage settings. 
On \textsc{CLINC-150}, \method\ achieves a mean of \textbf{93.65}\% (Known) and \textbf{95.88}\% (Unknown), exceeding strong boundary-based methods such as DA-ADB and ADB by sizable margins.
On \textsc{BANKING-77}, \method\ reaches \textbf{87.35}\% (Known) and \textbf{94.07}\% (Unknown), with the largest gains observed for Unknown detection. 
On \textsc{STACKOVERFLOW}, \method\ maintains \textbf{87.79}\% (Known) and \textbf{93.78}\% (Unknown), indicating robustness in a setting with many fine-grained classes. 
Across methods, confidence-, density-, and boundary-based baselines often exhibit a trade-off: improved Known performance coincides with degraded Unknown F1, or vice versa. In contrast, \method\ sustains high scores on both, consistent with its design objective.

\begin{table*}[!ht]
\centering
\caption{Macro F1 (\%) on Known and Unknown classes at label ratio $1.0$ under varying Known-intent coverage (25/50/75\%). Means are across the three coverage settings. Best results are \textbf{bold}.}
\label{Table:full-labeles-results}
\renewcommand{\arraystretch}{1.2}
\begin{tabular}{c c cc cc cc cc}
\toprule
\multirow{2}{*}{\textbf{Dataset}} & \multirow{2}{*}{\textbf{Method}} & \multicolumn{2}{c}{\textbf{25\%}} & \multicolumn{2}{c}{\textbf{50\%}} & \multicolumn{2}{c}{\textbf{75\%}} & \multicolumn{2}{c}{\textbf{Mean}} \\
\cmidrule(lr){3-4}\cmidrule(lr){5-6}\cmidrule(lr){7-8}\cmidrule(lr){9-10}
& & Known & Unknown & Known & Unknown & Known & Unknown & Known & Unknown \\
\midrule
\multirow{13}{*}{CLINC-150}
& (K+1)-way & 74.02 & 90.27 & 81.52 & 84.25 & 86.72 & 79.59 & 80.75 & 84.70 \\
& ADB & 77.85 & 92.36 & 85.12 & 88.60 & 88.97 & 84.85 & 83.98 & 88.60 \\
& ARPL & 73.01 & 89.63 & 80.87 & 81.81 & 86.10 & 74.67 & 80.00 & 82.04 \\
& DA-ADB & 79.57 & 93.20 & 85.58 & 90.10 & 88.43 & 86.00 & 84.53 & 89.77 \\
& DOC & 75.46 & 90.78 & 83.84 & 87.45 & 87.91 & 83.87 & 82.40 & 87.37 \\
& DeepUnk & 76.95 & 91.61 & 83.30 & 87.48 & 86.57 & 82.67 & 82.27 & 87.25 \\
& KNNCL & 78.85 & 93.56 & 83.25 & 87.85 & 86.14 & 82.05 & 82.75 & 87.82 \\
& LOF & 77.77 & 91.96 & 83.81 & 87.57 & 87.24 & 82.81 & 82.94 & 87.45 \\
& MDF & 49.43 & 84.89 & 61.60 & 62.31 & 72.21 & 51.33 & 61.08 & 66.18 \\
& MSP & 51.02 & 59.26 & 72.82 & 63.71 & 83.65 & 63.86 & 69.16 & 62.28 \\
& OpenMax & 73.74 & 90.69 & 80.59 & 85.50 & 86.38 & 80.44 & 80.24 & 85.54 \\
& SEG & 46.67 & 59.22 & 62.57 & 61.34 & 42.72 & 40.74 & 50.65 & 53.77 \\
\cmidrule{2-10}
& \textbf{DROID} & \textbf{93.98} & \textbf{98.59} & \textbf{93.76} & \textbf{96.52} & \textbf{93.20} & \textbf{92.52} & \textbf{93.65} & \textbf{95.88} \\
\midrule
\multirow{13}{*}{BANKING-77}
& (K+1)-way & 67.70 & 82.66 & 77.97 & 72.58 & 85.14 & 59.89 & 76.94 & 71.71 \\
& ADB & 70.92 & 85.05 & 81.39 & 79.43 & 86.44 & 67.34 & 79.58 & 77.27 \\
& ARPL & 62.99 & 83.39 & 77.93 & 71.79 & 85.58 & 61.26 & 75.50 & 72.15 \\
& DA-ADB & 73.05 & 86.57 & 82.54 & 81.93 & 85.93 & 69.37 & 80.51 & 79.29 \\
& DOC & 65.16 & 76.64 & 78.38 & 72.66 & 84.14 & 63.51 & 75.89 & 70.94 \\
& DeepUnk & 64.97 & 76.98 & 75.61 & 67.80 & 81.65 & 50.57 & 74.08 & 65.12 \\
& KNNCL & 65.54 & 79.34 & 75.16 & 67.21 & 81.76 & 51.42 & 74.15 & 65.99 \\
& LOF & 62.89 & 72.64 & 76.51 & 66.81 & 84.15 & 54.19 & 74.52 & 64.55 \\
& MDF & 44.80 & 85.70 & 64.27 & 57.72 & 75.47 & 33.43 & 61.51 & 58.95 \\
& MSP & 50.47 & 39.42 & 73.20 & 46.29 & 84.99 & 46.05 & 69.55 & 43.92 \\
& OpenMax & 53.42 & 48.52 & 75.16 & 55.03 & 85.50 & 53.02 & 71.36 & 52.19 \\
& SEG & 51.48 & 51.58 & 63.85 & 43.03 & 70.10 & 37.22 & 61.81 & 43.94 \\
\cmidrule{2-10}
& \textbf{DROID} & \textbf{85.04} & \textbf{96.63} & \textbf{87.89} & \textbf{94.38} & \textbf{89.11} & \textbf{91.21} & \textbf{87.35} & \textbf{94.07} \\
\midrule
\multirow{13}{*}{STACKOVERFLOW}
& (K+1)-way & 50.54 & 52.23 & 70.53 & 51.69 & 81.20 & 45.22 & 67.42 & 49.71 \\
& ADB & 77.62 & 90.96 & 85.32 & 87.70 & 86.91 & 74.10 & 83.28 & 84.25 \\
& ARPL & 60.55 & 72.95 & 78.26 & 73.97 & 85.24 & 62.99 & 74.68 & 69.97 \\
& DA-ADB & 80.87 & 92.65 & 86.71 & 88.86 & 87.66 & 74.55 & 85.08 & 85.35 \\
& DOC & 56.30 & 62.50 & 77.37 & 71.18 & 85.64 & 65.32 & 73.10 & 66.33 \\
& DeepUnk & 47.39 & 36.87 & 67.67 & 35.80 & 80.51 & 34.38 & 65.19 & 35.68 \\
& KNNCL & 41.79 & 15.26 & 61.50 & 8.50 & 76.16 & 7.19 & 59.82 & 10.32 \\
& LOF & 40.92 & 7.14 & 61.71 & 5.18 & 76.31 & 5.22 & 59.65 & 5.85 \\
& MDF & 48.13 & 83.03 & 62.60 & 50.19 & 73.96 & 28.52 & 61.56 & 53.91 \\
& MSP & 51.02 & 59.26 & 72.82 & 63.71 & 83.65 & 63.86 & 69.16 & 62.28 \\
& OpenMax & 47.51 & 34.52 & 69.88 & 46.11 & 82.98 & 49.69 & 66.79 & 43.44 \\
& SEG & 40.44 & 4.19 & 60.14 & 4.72 & 74.24 & 6.00 & 58.27 & 4.97 \\
\cmidrule{2-10}
& \textbf{DROID} & \textbf{87.72} & \textbf{97.22} & \textbf{87.83} & \textbf{93.99} & \textbf{87.81} & \textbf{90.13} & \textbf{87.79} & \textbf{93.78} \\
\bottomrule
\end{tabular}
\end{table*}

\subsection{Robustness to Limited Supervision}
\begin{figure*}[!t]
\centering
\includegraphics[width=\textwidth]{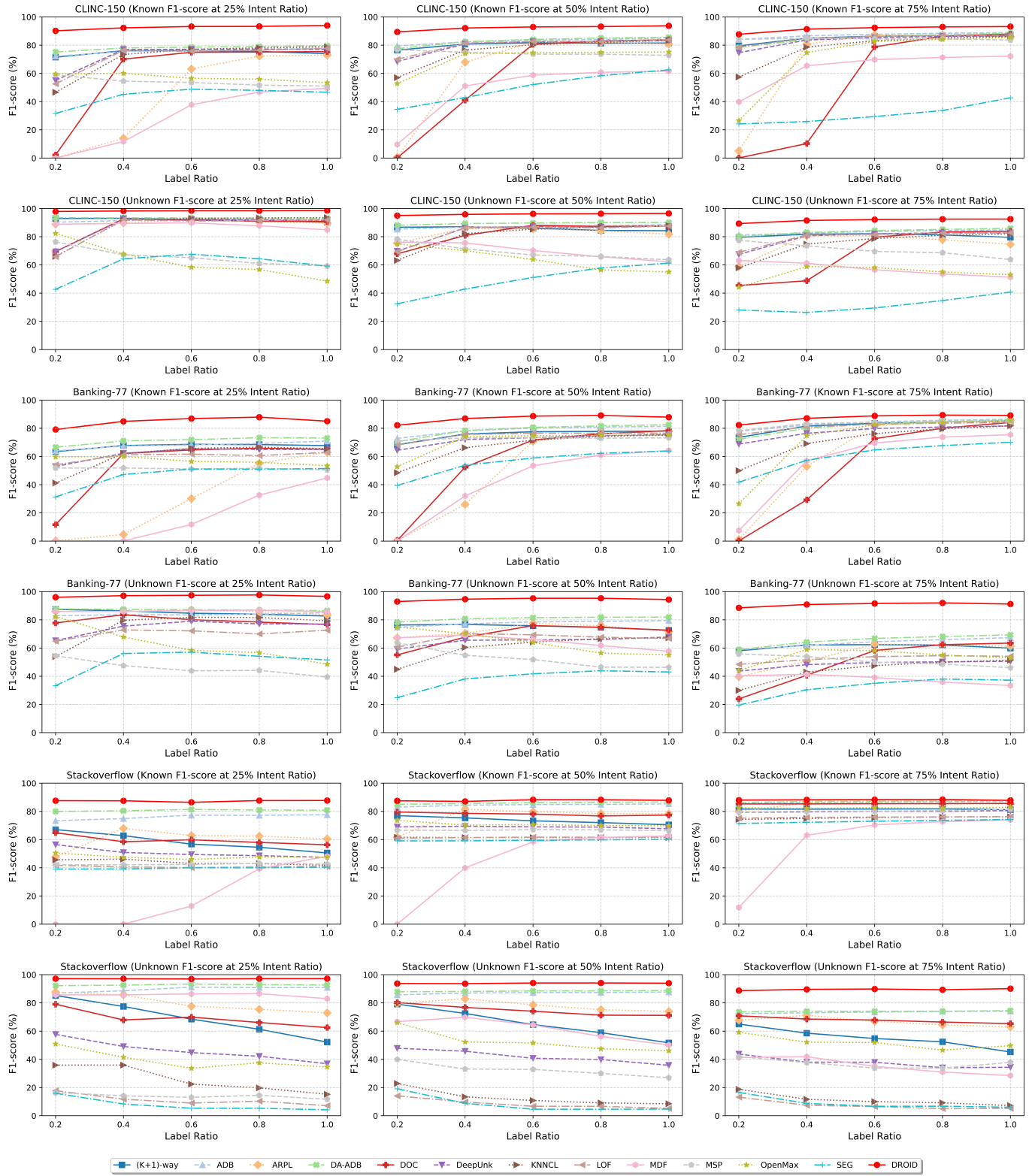}
\caption{Macro~F1 versus label ratio on \textsc{CLINC-150}, \textsc{BANKING-77}, and \textsc{STACKOVERFLOW}. Top: Known; Bottom: Unknown. Columns correspond to Known-intent coverage (25/50/75\%). \method\ (red) maintains high and balanced performance under limited supervision.}
\label{fig_FS_All}
\end{figure*}

Fig.~\ref{fig_FS_All} studies sensitivity to the label ratio ($\{0.2,0.4,0.6,0.8,1.0\}$) at each coverage level (25/50/75\%). \method\ remains strong even with scarce labels. For example, on \textsc{CLINC-150} at label ratio $0.2$, Known F1 is $\approx90\%$, whereas several baselines (e.g., ARPL, DOC, MDF) degrade severely under the same setting. As label availability increases, \method\ is either stable or improves slightly. Crucially, the method preserves balanced performance on Known and Unknown F1 across datasets, avoiding the pronounced trade-offs seen in confidence- and density-based alternatives.

\subsection{Effect of Threshold-Based Reclassification}
\begin{figure*}[!t]
\centering
\includegraphics[width=\textwidth]{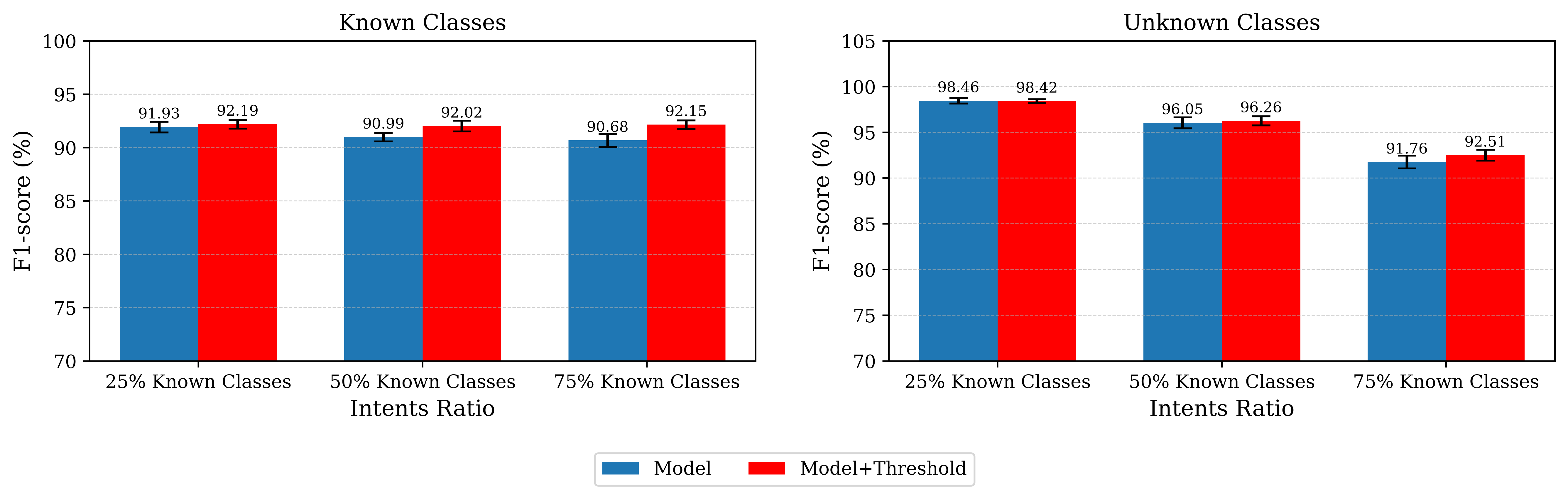}
\caption{Ablation on the thresholded decision rule (\textsc{CLINC-150}). Incorporating the calibrated threshold improves Unknown F1 at 25/50/75\% coverage while maintaining Known F1.}
\label{fig__threshold}
\end{figure*}

We quantify the contribution of the thresholded decision rule by ablating it on \textsc{CLINC-150} (Fig.~\ref{fig__threshold}). Adding the calibrated threshold consistently improves Unknown F1 at all coverage settings; Known F1 is preserved (around 91--92\%). Where error bars are shown, they indicate variability across runs. Similar behavior is observed on \textsc{BANKING-77} and \textsc{STACKOVERFLOW}. These findings support the utility of a simple, calibrated threshold for robust OOS rejection without harming in-domain accuracy.

\subsection{Impact of Class-Weighted Loss}
\begin{figure*}[!t]
\centering
\includegraphics[width=\textwidth]{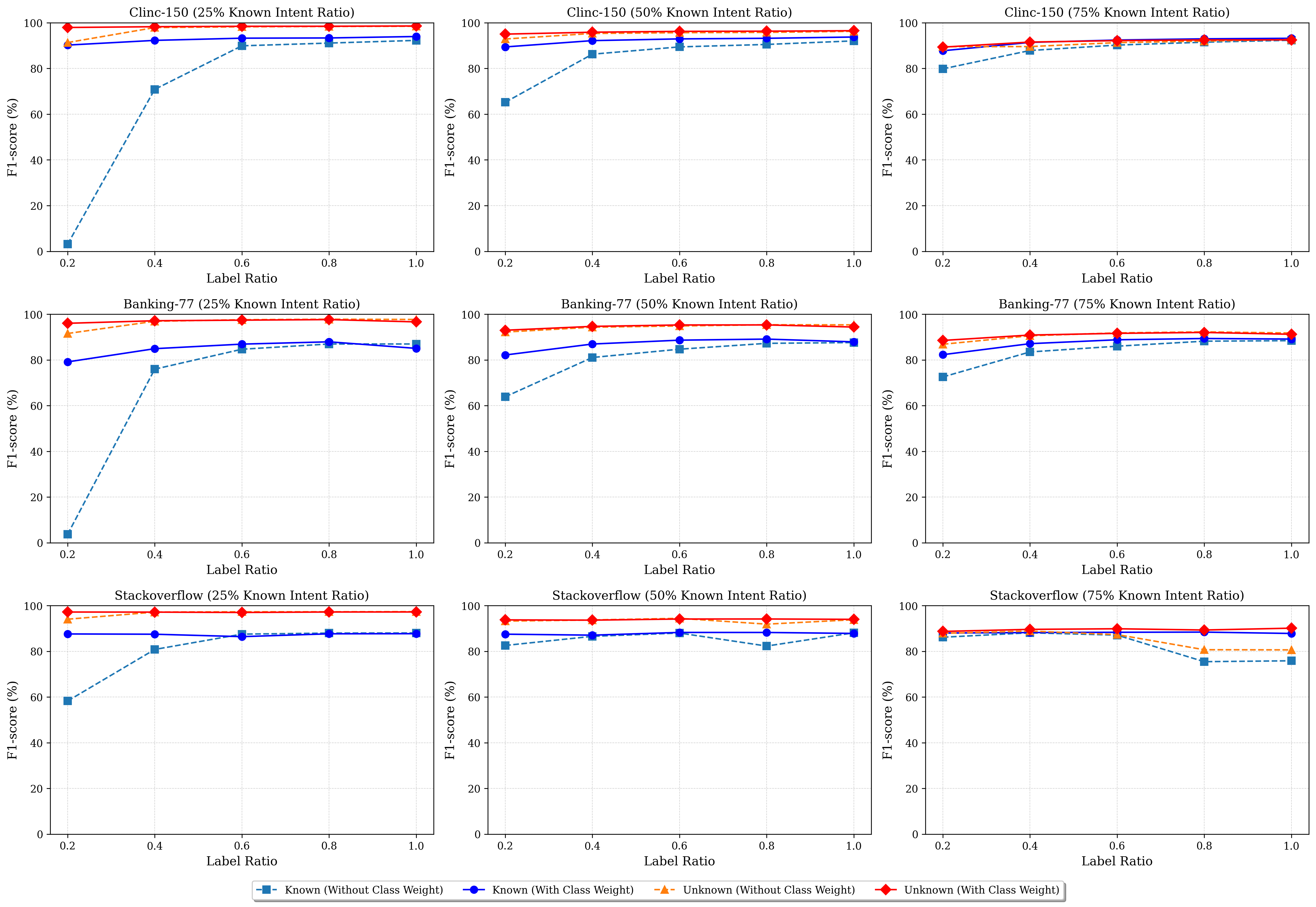}
\caption{Effect of class weighting across label ratios and coverages. Weights are most impactful for Known F1 in low-label regimes; Unknown F1 remains strong with marginal gains from weighting.}
\label{fig_class_weight}
\end{figure*}

We contrast training with and without class weights in Fig.~\ref{fig_class_weight}. Class weighting is particularly beneficial in extreme few-shot regimes: at low label ratios and low coverage, unweighted training may underfit Known classes, while weighting recovers strong Known F1. Unknown F1 is already high without weighting and benefits mainly from stabilization. As label ratio approaches $1.0$, the gap between weighted and unweighted settings narrows, indicating reduced imbalance sensitivity when supervision is ample.

\subsection{Effect of Outlier Quantity}
\begin{figure*}[!t]
\centering
\includegraphics[width=\textwidth]{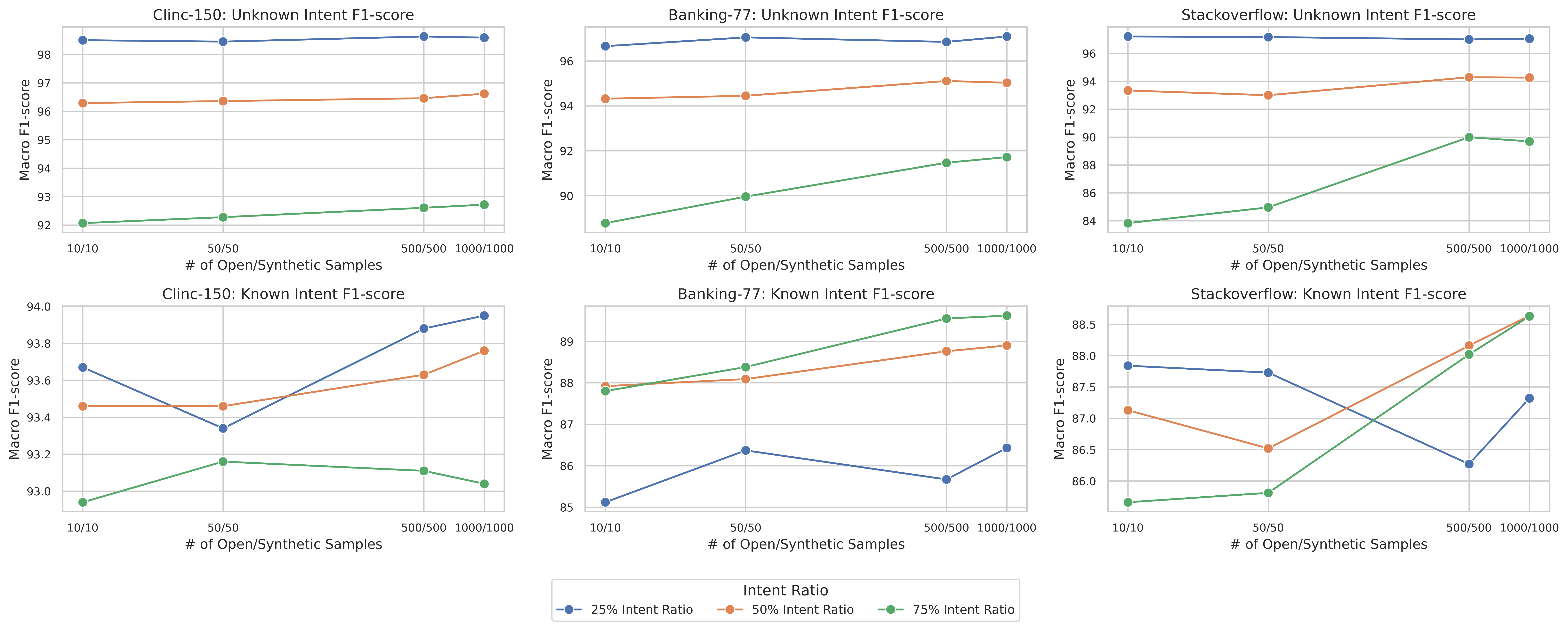}
\caption{Macro~F1 as a function of open-domain/synthetic outlier counts (rows: Unknown/ Known; columns: coverage 25/50/75\%). Adding diverse outliers modestly improves boundary sharpness and stabilizes results; gains saturate around a few hundred per type.}
\label{fig_open-synth-data}
\end{figure*}

Fig.~\ref{fig_open-synth-data} varies the count of open-domain and synthetic outliers from $10/10$ to $1000/1000$. Unknown F1 is highly robust, typically $95$--$98\%$ on \textsc{CLINC-150} and \textsc{BANKING-77} even with few outliers; on \textsc{STACKOVERFLOW}, Unknown F1 improves as outlier count increases, reflecting greater class-space complexity. Known F1 generally benefits or stabilizes with more outliers, suggesting tighter in-domain boundaries when the model is exposed to diverse negatives. Performance saturates around a few hundred per type, with no signs of overfitting at $1000/1000$.

\subsection{Effect of Encoding Strategy}
Table~\ref{Table:BANKING25-50-75} ablates sentence encoders on \textsc{BANKING-77}. The dual-encoder with in-domain TSDAE (\emph{TSDAE (CLINC-150)}$+$USE) dominates all alternatives across label ratios and coverages. Using a generic TSDAE (Roberta) or TSDAE adapted on unrelated corpora (AskUbuntu/SciDocs) yields intermediate performance; relying on USE alone is least effective. These results confirm that (i) unsupervised domain adaptation for TSDAE is critical, and (ii) dual encoding provides a complementary signal beyond a single encoder.

\begin{table*}[!ht]
\centering
\caption {Ablation on encoding strategies for \textsc{BANKING-77}: macro~F1 (\%) for Known/Unknown across coverages (25/50/75\%) and label ratios, under two outlier budgets (100/100 and 500/500).}
\label{Table:BANKING25-50-75}
\begin{tabular}{l c c c c c c c c c} %'l' for left-aligned,'c' for centered. Using'l' for first col.
\toprule % Top thick line

\textbf{Embedding} & \textbf{No. Open} & \textbf{No. Syn.} & \textbf{Label} & \multicolumn{6}{c}{\textbf{Intent Ratio}} \\
\cmidrule(lr){5-10} % Mid-rule for sub-headers, (lr) adds small space left/right
& \textbf{domain} & \textbf{} & \textbf{Ratio} & \multicolumn{2}{c}{\textbf{25\%}} & \multicolumn{2}{c}{\textbf{50\%}} & \multicolumn{2}{c}{\textbf{75\%}} \\
\cmidrule(lr){5-6} \cmidrule(lr){7-8} \cmidrule(lr){9-10} % Mid-rules for sub-sub-headers
& & & & \textbf{Known} & \textbf{Unk.} & \textbf{Known} & \textbf{Unk.} & \textbf{Known} & \textbf{Unk.} \\
\midrule % Standard mid-line

\multirow{10}{*}{TSDAE (Clinc-150) \& USE}
& \multirow{5}{*}{100} & \multirow{5}{*}{100} & 0.2 & 86.63 & 94.51 & 83.46 & 88.88 & 82.19 & 85.83 \\
& & & 0.4 & 88.48 & 94.41 & 86.90 & 89.97 & 85.39 & 86.21 \\
& & & 0.6 & 89.34 & 94.51 & 86.32 & 89.55 & 88.59 & 87.97 \\
& & & 0.8 & 87.53 & 93.98 & 87.96 & 89.39 & 89.04 & 88.13 \\
& & & 1.0 & 89.15 & 94.18 & 87.39 & 90.04 & 88.01 & 86.28 \\
\addlinespace % Adds a small vertical space for readability
& \multirow{5}{*}{500} & \multirow{5}{*}{500} & 0.2 & 80.15 & 92.76 & 79.76 & 88.34 & 79.39 & 87.60 \\
& & & 0.4 & 88.66 & 95.05 & 86.35 & 90.61 & 87.10 & 89.18 \\
& & & 0.6 & 90.59 & 95.50 & 88.26 & 92.18 & 87.65 & 89.22 \\
& & & 0.8 & 90.84 & 95.95 & 89.88 & 93.23 & 89.46 & 90.83 \\
& & & 1.0 & 90.82 & 95.55 & 90.33 & 92.87 & 90.07 & 90.49 \\
\midrule

\multirow{10}{*}{TSDAE(Roberta) \& USE}
& \multirow{5}{*}{100} & \multirow{5}{*}{100} & 0.2 & 57.73 & 84.61 & 65.53 & 81.02 & 69.54 & 76.02 \\
& & & 0.4 & 45.38 & 90.12 & 70.90 & 78.09 & 76.81 & 77.07 \\
& & & 0.6 & 60.97 & 84.14 & 68.90 & 71.06 & 81.93 & 78.14 \\
& & & 0.8 & 61.89 & 83.10 & 66.71 & 75.45 & 84.44 & 84.34 \\
& & & 1.0 & 67.00 & 86.57 & 67.65 & 76.33 & 85.21 & 84.41 \\
\addlinespace
& \multirow{5}{*}{500} & \multirow{5}{*}{500} & 0.2 & 56.31 & 86.77 & 66.05 & 80.08 & 76.40 & 80.78 \\
& & & 0.4 & 59.69 & 88.07 & 66.01 & 77.96 & 82.30 & 82.25 \\
& & & 0.6 & 58.94 & 85.82 & 69.94 & 80.53 & 83.22 & 82.33 \\
& & & 0.8 & 47.09 & 90.43 & 70.22 & 80.91 & 83.81 & 82.67 \\
& & & 1.0 & 65.61 & 89.84 & 70.77 & 78.76 & 84.47 & 83.83 \\
\midrule

\multirow{10}{*}{TSDAE (Askubuntu) \& USE}
& \multirow{5}{*}{100} & \multirow{5}{*}{100} & 0.2 & 62.34 & 88.89 & 70.40 & 85.27 & 77.11 & 79.66 \\
& & & 0.4 & 64.19 & 87.98 & 74.40 & 83.69 & 80.49 & 80.36 \\
& & & 0.6 & 65.02 & 86.21 & 75.33 & 84.48 & 82.57 & 83.39 \\
& & & 0.8 & 64.01 & 85.66 & 74.44 & 82.38 & 82.59 & 81.92 \\
& & & 1.0 & 63.40 & 86.57 & 74.33 & 83.75 & 85.27 & 84.93 \\
\addlinespace
& \multirow{5}{*}{500} & \multirow{5}{*}{500} & 0.2 & 59.85 & 90.04 & 66.82 & 84.12 & 74.62 & 80.27 \\
& & & 0.4 & 64.64 & 90.16 & 73.01 & 82.71 & 81.05 & 81.96 \\
& & & 0.6 & 64.51 & 89.95 & 74.96 & 83.73 & 82.15 & 82.29 \\
& & & 0.8 & 64.32 & 89.51 & 74.24 & 84.06 & 84.32 & 83.35 \\
& & & 1.0 & 62.43 & 88.52 & 76.23 & 84.61 & 81.77 & 82.09 \\
\midrule

\multirow{10}{*}{TSDAE (Scidocs) \& USE}
& \multirow{5}{*}{100} & \multirow{5}{*}{100} & 0.2 & 62.18 & 89.96 & 69.10 & 83.10 & 74.75 & 79.13 \\
& & & 0.4 & 61.61 & 89.69 & 71.43 & 82.11 & 79.44 & 80.36 \\
& & & 0.6 & 62.20 & 89.30 & 72.73 & 83.45 & 79.47 & 79.56 \\
& & & 0.8 & 56.39 & 89.03 & 75.03 & 82.77 & 82.55 & 80.66 \\
& & & 1.0 & 60.25 & 87.83 & 72.12 & 80.63 & 80.96 & 82.73 \\
\addlinespace
& \multirow{5}{*}{500} & \multirow{5}{*}{500} & 0.2 & 57.58 & 90.30 & 66.62 & 83.40 & 73.56 & 79.59 \\
& & & 0.4 & 63.67 & 90.34 & 71.55 & 84.30 & 81.21 & 82.02 \\
& & & 0.6 & 60.88 & 89.05 & 69.61 & 82.74 & 80.06 & 80.92 \\
& & & 0.8 & 66.22 & 89.14 & 74.87 & 84.00 & 82.97 & 82.12 \\
& & & 1.0 & 56.32 & 88.39 & 74.77 & 84.53 & 82.17 & 81.45 \\
\midrule

\multirow{10}{*}{USE Only}
& \multirow{5}{*}{100} & \multirow{5}{*}{100} & 0.2 & 57.69 & 87.31 & 64.26 & 83.18 & 75.41 & 79.41 \\
& & & 0.4 & 60.84 & 83.54 & 68.55 & 79.24 & 64.06 & 75.76 \\
& & & 0.6 & 61.86 & 85.10 & 71.21 & 77.20 & 78.26 & 75.22 \\
& & & 0.8 & 63.61 & 84.21 & 66.09 & 83.90 & 81.60 & 82.71 \\
& & & 1.0 & 64.12 & 85.42 & 69.11 & 80.65 & 82.93 & 78.12 \\
\addlinespace
& \multirow{5}{*}{500} & \multirow{5}{*}{500} & 0.2 & 58.05 & 86.42 & 63.38 & 81.44 & 76.57 & 81.02 \\
& & & 0.4 & 60.85 & 84.71 & 61.79 & 83.19 & 78.98 & 80.14 \\
& & & 0.6 & 58.76 & 88.82 & 72.75 & 79.96 & 68.60 & 76.52 \\
& & & 0.8 & 63.64 & 86.90 & 75.80 & 84.11 & 79.57 & 80.11 \\
& & & 1.0 & 52.74 & 90.00 & 76.55 & 85.65 & 81.09 & 77.98 \\
\bottomrule % Bottom thick line
\end{tabular}
\end{table*}

Across datasets, coverage regimes, and label budgets, \method\ consistently delivers state-of-the-art macro~F1 on Known and Unknown classes. Its calibrated threshold enhances OOS rejection without harming in-domain accuracy; class weighting is crucial under extreme few-shot conditions; diverse outliers modestly sharpen boundaries; and the dual-encoder with in-domain TSDAE is a key contributor to overall gains.

\section{Discussion}
\label{sec:discussion}

This section reflects on the empirical findings and design choices of \textbf{DROID}, relates them to the literature, and outlines limitations and future directions. We are emphasizing generalization, efficiency, and deployability rather than repeating numerical results already presented in Section~\ref{sec:results}.

\subsection{Summary of Findings and Generalization}
Across three dialogue benchmarks---\textsc{CLINC-150}, \textsc{BANKING77}, and \textsc{STACKOVERFLOW}---DROID delivers consistently strong macro-F1 on both in-domain (known) and out-of-scope (OOS) intents ((Table~\ref{Table:full-labeles-results}). Importantly, these gains persist under varying known-intent proportions (25\%, 50\%, 75\%) and reduced label ratios (Fig.~\ref{fig_FS_All}), indicating that the method is robust to (i) incomplete intent coverage and (ii) limited supervision. The absence of a marked trade-off between known and OOS performance (Section~\ref{sec:results}) suggests that the learned representation and decision rule are well-calibrated for open-intent settings.

\subsection{Dual Representations and a Calibrated Threshold}
DROID’s design integrates \emph{two complementary encoders}---a general-purpose USE branch and a domain-adapted TSDAE branch---fused by a lightweight head (Section~\ref{sec:droid}). This pairing balances broad semantic coverage with domain-sensitive nuance, improving cluster compactness and inter-class separability in the embedding space compared to single-encoder baselines. A single calibrated threshold—tuned on ID validation (Section~\ref{sec:experimental_setup})—implements a transparent inference-time rejection rule without post-hoc modules or parametric tail assumptions, achieving strong OOS recall while remaining simple and interpretable, unlike methods that add post-hoc scoring models or adversarial objectives \cite{xu-etal-2021-unsupervised,Zhang_2021,DA-ADB_2023}. Ablations in Fig.~\ref{fig__threshold} show that thresholding materially improves OOS recognition without eroding ID performance, aligning with the intuition that confidence-aware rejection is an effective open-set primitive in text classification.

\subsection{Role of Outlier Data: Synthetic vs.\ Open-Domain}
Training with a mixture of \emph{synthetic feature-space outliers} and \emph{open-domain negatives} (Section~\ref{sec:experimental_setup}) follows the spirit of \cite{Zhan_2021} while embedding it in a dual-representation pipeline. Our analyses (Fig.~\ref{fig_open-synth-data}) shows that (i) moderate quantities of both sources suffices; (ii) synthetic outliers are particularly effective for tightening decision regions around known intents (benefiting both kown and OOS macro-F1); and (iii) returns saturate as counts approach several hundred per type, suggesting diminishing returns beyond a few hundred diverse samples. Practically, this implies that \emph{in-domain-informed} synthetic outliers are often more valuable than large volumes of unrelated text, helping populate the boundary region where confusions are most likely.

\subsection{Efficiency and Deployability}
DROID’s trainable footprint is $\mathbf{1{,}559{,}808}$ parameters (Section~\ref{sec:droid}), orders of magnitude smaller than many open-intent pipelines relying on full Transformer fine-tuning \cite{Zhan_2021}. The frozen encoders, lightweight heads, and single-threshold rule reduce training complexity and inference latency. This balance of accuracy and efficiency is salient for latency-sensitive dialogue systems and resource-constrained settings (edge or on-device). In contrast to LLM-based open-intent baselines (Section~\ref{sec:related_literature}), DROID attains competitive accuracy without incurring heavy memory or serving costs.

\subsection{Limitations and Future Directions}
\textbf{(i) Scope of evaluation.} Experiments cover single-turn English utterances on three benchmarks; extending to multi-turn settings, multilingual corpora, and domain drift scenarios is a priority. Given TSDAE’s unsupervised adaptation, multilingual or domain-specific TSDAE pretraining is a natural path (cf.\ Section~\ref{sec:experimental_setup}).

\textbf{(ii) Static thresholding.} The calibrated threshold is global and fixed per setting. While Fig.~\ref{fig__threshold} shows strong utility, \emph{context-aware} or \emph{adaptive} thresholding (e.g., conditioned on utterance uncertainty or class priors) could further stabilize OOS rejection under shift.

\textbf{(iii) Encoder consolidation.} Dual encoders yield complementary gains (Table~\ref{Table:BANKING25-50-75}), but maintaining two branches increases memory compared to a single encoder. Future work could explore knowledge distillation from the dual-branch model into a unified encoder while preserving open-set separability.

\textbf{(iv) Outlier generation.} Our synthetic outliers are convex combinations \emph{post-encoding}. More expressive generators (e.g., learned feature perturbations or text-level generators constrained by semantic similarity) may populate decision boundaries more effectively while controlling for bias.

\subsection{Positioning within the Literature}
DROID’s contributions sit between open-set decision rules and representation learning advances reviewed in Section~\ref{sec:related_literature}: it eschews adversarial/boundary-heavy training \cite{Zhang_2021,DA-ADB_2023} and separate post-hoc detectors \cite{xu-etal-2021-unsupervised} in favor of (i) enriched sentence-level representations (USE+TSDAE) and (ii) a single calibrated threshold within a unified \((K{+}1)\) classifier. This combination yields state-of-the-art results (Table~\ref{Table:full-labeles-results}) with a simpler deployment path.

\section{Conclusion}
\label{sec:conclusion}

This work presented \textbf{DROID}, a compact dual-encoder framework for robust out-of-scope (OOS) intent detection. By combining a general-purpose semantic encoder (USE) with a domain-adapted denoising autoencoder (TSDAE), DROID learns complementary representations that enhance both in-domain discrimination and out-of-domain rejection. The integration of a calibrated, threshold-baswhich shareassification mechanism further improves reliability without the lead toy of adversarial or post-hocreliably  modeling. Empirical analss \textsc{CLINC-150}, \textsc{BANKING77}, and \textsc{STACKOVERFLOW} confirm consistent performance gains and stability under limited supervision.

Beyond empirical results, DROID highlights a broader principle: coupling heterogeneous encoders with calibrated confidence estimation offers an efficient pathway toward open-world intent understanding. The results suggest that representational diversity and explicit decision calibration can jointly improve robustness, even in lightweight architectures. This insight may inform future work on confidence-aware neural classifiers more generally.

The model’s efficiency—only 1.5M trainable parameters—demonstrates that strong OOS detection does not require large-scale fine-tuning or complex objectives, supporting deployment in real-time or resource-constrained dialogue systems. Future extensions will explore multilingual and continual adaptation, aiming to extend the DROID design to evolving, multi-lingual intent spaces. Investigating theoretical properties of dual-encoder calibration and integrating explainability for human-in-the-loop intent verification remain promising avenues for advancing trustworthy conversational AI.

\section{Acknowledgments}
This project has received funding from Enterprise Ireland and the European Union's Horizon 2020 Research and Innovation Programme under the Marie Skłodowska-Curie grant agreement No 847402.

\bibliographystyle{IEEEtran}
\bibliography{IEEEabrv,lrec-coling2024-example}

\end{document}